%% file: main.tex
\documentclass[10pt,twocolumn,letterpaper]{article}

\usepackage{iccv}              


\definecolor{iccvblue}{rgb}{0.21,0.49,0.74}
\usepackage[pagebackref,breaklinks,colorlinks,allcolors=iccvblue]{hyperref}

\usepackage{amsmath}
\usepackage{amssymb}
\usepackage{booktabs}

\usepackage{algorithm}
\usepackage{algorithmicx}
\usepackage[noend]{algpseudocode}

\usepackage{graphicx} 
\usepackage{amsmath}
\usepackage{amssymb}
\usepackage{booktabs}
\usepackage{xcolor}

\def\eg{\emph{e.g}\onedot} 
\def\ie{\emph{i.e}\onedot}

\def\etal{\emph{et al}\onedot}
\makeatother

\usepackage{multirow}
\usepackage{diagbox}
\usepackage[capitalize]{cleveref}
\crefname{section}{Sec.}{Secs.}
\crefname{section}{Section}{Sections}
\crefname{table}{Table}{Tables}

\def\X{{\mathbf X}}
\def\G{{\mathbf G}}
\def\N{{\mathbf N}}
\def\AE{{\mathbf {\hat X}}}

\def\F{{f_{\theta}}}

\def\paperID{XXX} 
\def\confName{CV}
\def\confYear{2025}

\title{Towards Calibration Enhanced Network by Inverse Adversarial Attack}

\author{
Yupeng, Cheng\\
\textit{Nanyang Technological University}\\
Singapore \\
{\tt\small yupeng.cheng@ntu.edu.sg}
\and
Zi Pong, Lim\\
\textit{Continental Corporation}\\
Singapore \\
{\tt\small zi.pong.lim@continental-corporation.com}
\and
Sarthak Ketanbhai, Modi\\
\textit{Nanyang Technological University}\\
Singapore \\
{\tt\small sarthak005@e.ntu.edu.sg}
\and
Yon Shin, Teo\\
\textit{Continental Corporation}\\
Singapore \\
{\tt\small yon.shin.teo@continental-corporation.com}
\and
Yushi, Cao\\
\textit{Nanyang Technological University}\\
Singapore \\
{\tt\small yushi002@e.ntu.edu.sg}
\and
Shang-Wei, Lin\\
\textit{Nanyang Technological University}\\
Singapore \\
{\tt\small shang-wei.lin@ntu.edu.sg}
}

\newcommand{\yupeng}[1]{\textbf{\textcolor{red}{Yupeng: #1}}}

\begin{document}
\maketitle
\input{sec/0_abstract}    
\input{sec/1_intro}
\input{sec/2_related_work}
\input{sec/3_method}

\input{sec/4_experiments}
\input{sec/5_conclusion}

{
    \small
    \bibliographystyle{IEEEtran}
    \bibliography{main}
}

\end{document}

%% file: sec/0_abstract.tex
\begin{abstract}
   Model calibration has been a focus of study for many years, as a well-calibrated model is crucial for reliable predictions in safety-critical tasks, such as face authentication and autonomous driving systems. While adversarial examples can reveal overconfidence issues in models under slight perturbations, the opposite issue—underconfidence—has rarely been addressed. In this paper, we introduce a novel task, termed Inverse Adversarial Attack (IAA), to highlight potential underconfidence issues in models. Additionally, we incorporate examples generated by IAA during the training process to improve model calibration. Unlike traditional adversarial training, which primarily addresses overconfident adversarial examples, inverse adversarial training focuses on adjusting underconfident cases. Ultimately, by combining adversarial and inverse adversarial training, we can develop a model that excels in both robustness and calibration.
\end{abstract}






%% file: sec/1_intro.tex
\section{Introduction}
\label{sec:intro}

\begin{figure}[t]
\centering
        \centering
        \includegraphics[width=1\linewidth]{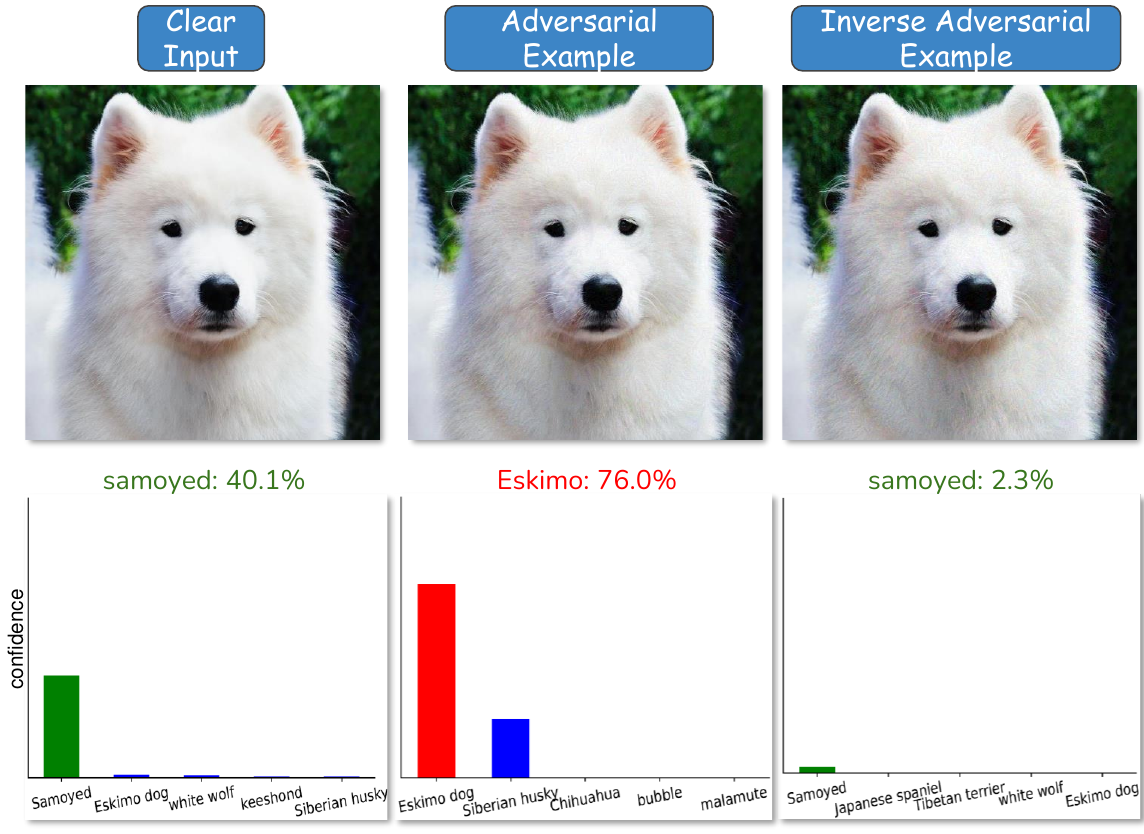}
        \caption{An image of a \textbf{Samoyed} dog from the ImageNet dataset, predicted by a pretrained ResNet50 DNN classifier, demonstrates the presence of the Inverse Adversarial Example. (a) The clear input is correctly classified as \textbf{Samoyed} with a confidence of $40.1\%$. (b) The Adversarial Example, with a perturbation of $\epsilon=0.03$, is misclassified as \textbf{Eskimo} with a confidence of $76\%$. (c) The Inverse Adversarial Example, also with a perturbation of $\epsilon=0.03$, is correctly classified as \textbf{Samoyed}, but with a very low confidence of $2.3\%$. The prediction confidence distributions for the corresponding inputs are shown at the bottom.}
\label{IAA:fig:cover}
\end{figure}



Calibration is a critical property for ensuring reliable prediction in safety-related tasks, \eg, face recognition, commercial judgment \cite{buchel2022deep,aastebro2007calibration}, diagnosis \cite{gallaher2004impact,caruana2015intelligible} and self-driving \cite{wu2021way,bojarski2016end}. 
In many real-world tasks, accuracy alone is not the only measure of model quality, as the true label of an input case is not known in advance. Therefore, confidence is also considered alongside the model's output to determine how to proceed with decisions and whether additional procedures are needed \cite{wang2023calibration}. Deploying uncalibrated models can be risky: if the output is overconfident, it is likely to mislead subsequent decisions. Conversely, if the output is underconfident, the system may cast unnecessary doubt, leading to excessive interventions and higher costs due to expert involvement, ultimately reducing efficiency.

Adversarial Attacks (AA) typically generate examples that cause networks to become overconfident in their predictions \cite{teixeira2023reducing}. Thus, adversarial training (AT) for robustness enhancement can partially address the overconfidence issue by integrating adversarial examples (AEs) into the training process, as these examples often exhibit high confidence despite incorrect predictions.
Some studies even incorporate prediction confidence as a term in the loss function to address the overconfidence of AEs \cite{stutz2020confidence}.

However, since miscalibration involves both overconfidence and underconfidence, adversarial training primarily focuses on correcting overconfident cases (incorrect to correct predictions) while neglecting underconfident cases. 
In contrast, this study introduces a novel task that generates examples correctly predicted but with limited confidence, termed ``Inverse Adversarial Attack'' (IAA). The corresponding examples are ``Inverse Adversarial Examples'' (IAEs). They can thus be leveraged to expose underconfidence issues into training process, termed Inverse Adversarial Training (IAT), and further enhance model calibration against tiny perturbations.

An example from the ImageNet dataset \cite{deng2009imagenet} is presented in \cref{IAA:fig:cover}. It is evident that AE produces an incorrect prediction with a confidence level even higher than that of the clear input, highlighting the overconfidence issue in the target model. On the other hand, IAE provides the correct prediction but with very limited confidence ($2.3\%$), exposing the underconfidence problem in a well-trained model.

Generally, our contributions are as follow:
\begin{itemize}
\item We propose a novel task, \ie, Inverse Adversarial Attack, which focuses on generating underconfidence examples, referred to as Inverse Adversarial Examples. These examples satisfy two conditions: they are \textbf{correctly predicted} by the model but with \textbf{limited confidence}.
\item We introduce two cross-entropy loss terms to represent the two conditions of underconfidence examples, and combine them using a positive weight parameter to construct the final loss function $\mathcal{L}$. And we also mathematically prove that, by minimizing the loss function we can ensure the IAEs can truely satisfy the two conditions of underconfidence examples.
\item Inspired by Adversarial Training, we introduce the IAEs as an augmentation in training process, and propose Inverse Adversarial Training which aim to defend the underconfidence attack.
\item The experiments demonstrate the effectiveness of IAA in exposing a model's underconfidence issues, while IAT effectively defends against such attacks. The combination of AT and IAT is shown to be practical, providing a model that is both robust and well-calibrated.
\end{itemize}


%% file: sec/2_related_work.tex
\section{Related Works}
\label{sec:relatedwork}
\subsection{Calibration Methods}

%

Generally, model calibration can be achieved through two methods: post-hoc calibration and calibration training.
Post-hoc calibration focuses on adjusting a model's output confidence as a post-processing step. Its main advantage is that it does not require updating or retraining the existing model. Zadrozny \etal \cite{zadrozny2001obtaining} first introduced histogram binning to produce calibrated probability estimates for decision tree and naive Bayesian classifiers. Subsequently, they proposed isotonic regression as a calibration technique \cite{zadrozny2002transforming}. Additionally, a more efficient and cost-effective post-hoc method is Platt Scaling \cite{platt1999probabilistic}, along with its single-parameter extension, Temperature Scaling, which has inspired various extensions \cite{mozafari2018attended,kull2019beyond,ding2021local}.
In terms of calibration training methods, Zhang \etal \cite{zhang2017mixup} proposed mixup regularization to create simple linear relationships between training examples. Tomani \etal \cite{tomani2021towards} introduced an entropy-encouraging loss term combined with an adversarial calibration loss term to develop well-calibrated and trustworthy models.

\subsection{Adversarial Attack}


%

There are two main settings for adversarial attacks. The first is the black-box attack, where the attacker has limited information about the target models \cite{carlini2017towards}. The second is the white-box attack, in which the attacker has full access to the target models \cite{goodfellow2014explaining}.
One widely recognized defense method is adversarial training. Madry \etal \cite{madry2017towards} employed adversarial examples as data augmentation during training to improve model robustness. Stutz \etal \cite{stutz2020confidence} introduced Confidence-Calibrated Adversarial Training (CCAT), which minimizes the confidence of adversarial examples during training to provide signals for identifying attack cases. Combined with the low accuracy of adversarial examples, this approach produces a highly calibrated model against perturbations. Similarly, Chen \etal \cite{chen2024integrating} proposed incorporating an adversarial-calibrated entropy term during training, which limits the confidence of perturbed inputs.

Both CCAT and ACE address calibration issues during training by incorporating additional terms. However, their focus is primarily on the overconfidence problem, often neglecting the issue of underconfidence.

UAC pointing out the important of calibration of a model. it utilize the function to increase the confidence to generate UAC. however, they only decrease the confidence, but do not try to change the prediction. in that case the model with low accuracy can not 



%% file: sec/3_method.tex
\section{Method}
\label{IAT:sec:method}
In this section, we provide a detailed introduction to our Inverse Adversarial Attack, outlining the design of the loss function to meet the two conditions for underconfidence examples: \textbf{correct prediction} and \textbf{limited confidence}. Furthermore, we elaborate on the framework of the defense method, \ie, Inverse Adversarial Training.

\subsection{Inverse Adversarial Attack}

A perturbation $\N$ is inevitable during the storage or transfer of multimedia data. Consequently, any input image, $\hat{\X}$, can be regarded as a combination of the clear input $\X$ and the perturbation $\N$:
\begin{align}\label{IAT:eq:addperturb}
\AE = \X + \N.         
\end{align}



The objective of IAA is to generate specific noise $\N$, which causes the model $\F(\cdot)$ to output the lowest maximum confidence while making a correct prediction:



\begin{equation}
\label{IAT:eq:IAE_define} 
\begin{aligned}
N = 
&\mathop{\mathrm{argmin}}\limits_{N} (max(\F(\X+\N))), \\
& s.t.~~ Label(\F(\X+\N)) = \G \\
& s.t.~~\|\N\|_{\infty} \leq \epsilon,
\end{aligned}
\end{equation}
where $Label(\cdot)$ means the predicted label of input. 
Since this objective (under-confidence) is opposite to adversarial attacks, we refer to it as \textit{Inverse Adversarial Attack} (IAA).
To jointly optimize the condition and object, we utilize the loss function for recognition, \eg, cross entropy loss, to replace $Label(\cdot)$ and $\mathop{\mathrm{argmin}}\limits_{N}(\cdot)$. Thus, the \cref{IAT:eq:IAE_define} is modified to:

\begin{equation}
\begin{aligned}
\mathcal{L}(\theta, &\X, \N, \G) =\\
&\mathcal{L}_{CE}(\F(\X+\N), \G) +  \lambda * \mathcal{L}_{CE}(\F(\X+\N), 1/K )
\end{aligned}
\label{IAT:eq:IAE2}
\end{equation}


where $\mathcal{L}_{CE}(\cdot, \cdot)$ refers to cross entropy loss.  $\epsilon$ is the maximum perturbation. The first term guarantees a correct prediction result and the second term affects the prediction confidence. $\lambda$ controls the weight between two conditions.
By minimizes the $\mathcal{L}$, IAA provides the underconfident Inverse Adversarial Examples (IAEs):

\begin{align}
\label{IAT:eq:N-IAE}
N_{\mathrm{IAE}} = &\mathop\mathrm{argmin}\limits_{N} \mathcal{L}
\end{align}

\begin{figure}[t]
\centering
        \centering
        \includegraphics[width=1\linewidth]{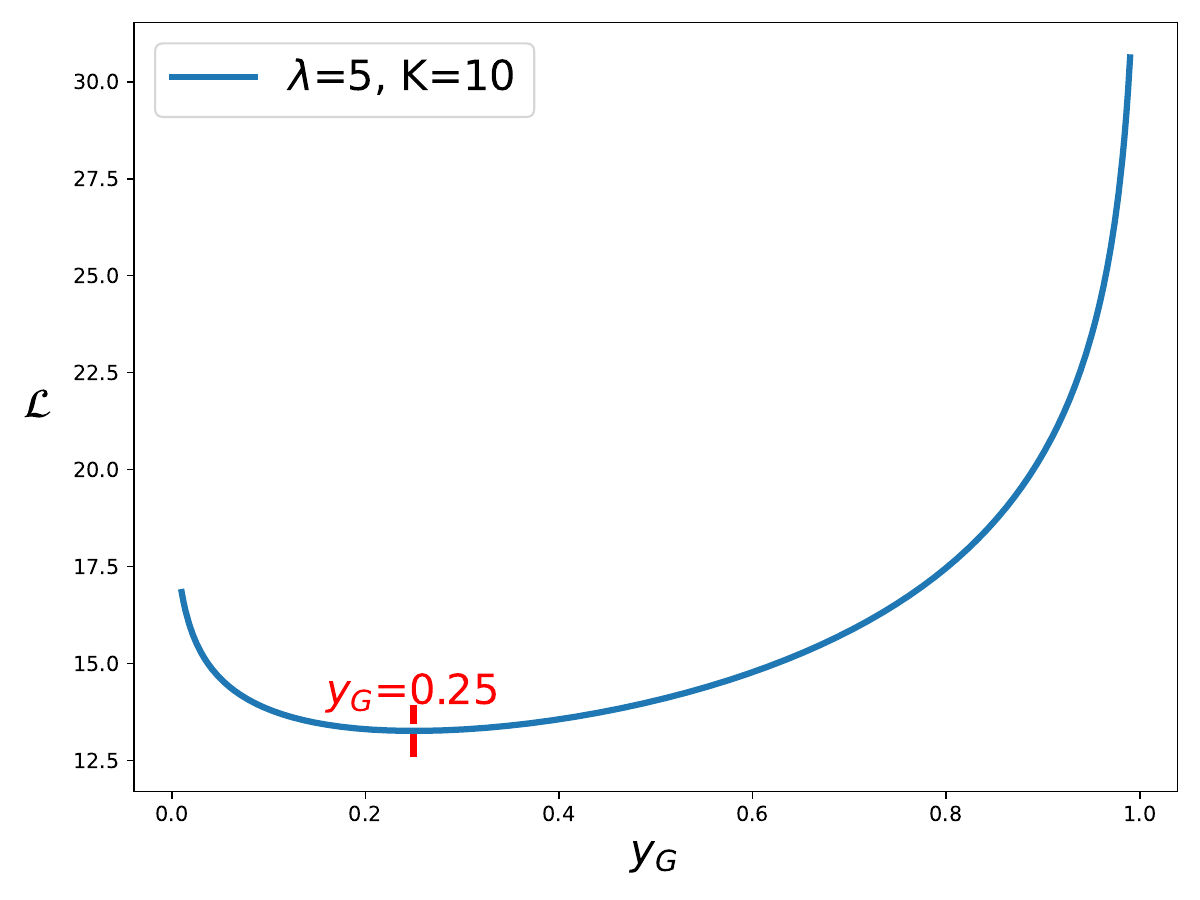}
        \caption{The analysis of $\mathcal{L}$ in \cref{IAA:eq:optim} with $\lambda=5,~K=10$. $\mathcal{L}$ reaches its minimum when $y_G = 0.25$.}    
\label{IAA:fig:fp}
\end{figure}




Generally, we solve \cref{IAT:eq:N-IAE} by sign gradient descent. Note the number of attack iteration as $T$, and each pixel value in $\N$ is updated for every iteration $t$:

\begin{align}\label{IAA:eq:advupdate}
\N_{t} = \N_{t-1}~ - ~ \alpha~ \mathrm{sign}(\nabla \mathcal{L}_{\N_{t-1}}),
\end{align}
where $\alpha$ is the step size of each iteration. $\mathrm{sign}(\cdot)$ represents the signum function. $\nabla \mathcal{L}_{\N_{t-1}}$ denotes the gradient of $\N_{t-1}$ with respect to the objective function.


\subsection{Optimization Analysis}
\label{IAA:subsec:optim}
Taken an input image, we note $y_k$ as the predict confidence of class $k$ along with classifier $\F$ and the entire predict distribution is $\mathbf{y} = \{y_1, y_2,...,y_K\}$. $K$ is the number of classes.
Let \( G \) be the ground truth label, and \( \mathbf{\hat{y}} \) be the one-hot representation of the ground truth, defined as:
\begin{equation}
\label{IAA:eq:y_hat}
\mathbf{\hat{y}} = 
\begin{cases}
\hat{y}_k = 1, & \text{if } k = G; \\
\hat{y}_k = 0, & \text{otherwise}.
\end{cases}
\end{equation}
Usually, the constrain of the $\sum_{k=1}^{K} y_k = 1$ is guaranteed by the final softmax layer $s(\cdot)$. 
Let's note $\mathbf{z}=\{z_k,k=1,2,...,K\}$ as the output logits before the final softmax layer, \ie, $ \mathbf{y} = s(\mathbf{z})$.
Moreover, we denote the uniform distribution $\mathbf{y^U}$ as $\{y^U_k = 1/K, i=1,2,...,K\}$.
Then, according to the definition of cross entropy, the loss function \cref{IAT:eq:IAE2} is reformulated as:

\begin{equation}
\begin{aligned}\label{IAA:eq:optim}
\mathcal{L} &= \mathcal{L}_{CE}(\mathbf{y},\mathbf{\hat{y}}) + \lambda * \mathcal{L}_{CE}(\mathbf{y},\mathbf{y^U}),
\end{aligned}
\end{equation}

where
As $z_k$ can be treated as individual, we then differentiate $\mathcal{L}$ with respect to $z_k$:
\begin{equation}
\begin{aligned}\label{IAA:eq:optim_dif}
\frac{\partial \mathcal{L}}{\partial z_k} 
&= \frac{\mathcal{L}_{CE}(\mathbf{y},\mathbf{\hat{y}})}{\partial z_k} +  \lambda * \frac{\mathcal{L}_{CE}(\mathbf{y},\mathbf{y^U})}{\partial z_k}\\
&= y_k-\hat{y}_k + \lambda * (y_k - y^U_k)\\
&= (1+\lambda)*y_k - (\hat{y}_k + \lambda * y^U_k)
\end{aligned}
\end{equation}

As $\lambda>0$, $1+\lambda>0$,  indicating that $\mathcal{L}$ is a concave function. Moreover, its minimum is attained under the condition:
\begin{equation}
\begin{aligned}\label{IAA:eq:optim_dif_k}
\frac{\partial L}{\partial z_k} &=0\\
(1+\lambda)*y_k - (\hat{y}_k + \lambda * y^U_k)&=0\\
y_k &=\frac{\hat{y}_k + \lambda * y^U_k}{1+\lambda}\\
\end{aligned}
\end{equation}

Finally, substituting \cref{IAA:eq:y_hat} yields:
\begin{equation}
\label{IAA:eq:optim_yk}
y_k = 
\begin{cases}
\dfrac{1 + \lambda / K}{1+\lambda}, & \text{if } k = G; \\
\dfrac{\lambda / K}{1+\lambda}, & \text{otherwise}.
\end{cases}
\end{equation}


In \cref{IAA:fig:fp}, we generate a curve to better illustrate the relationship between $\mathcal{L}$ and the predicted probability $y_{G}$, along with the model $\F$, when $\lambda = 5$ and $K=10$. The figure shows that the loss $\mathcal{L}$ reaches its minimum at $y_G = 0.25$, which is consistent with the previous analysis.

As $y_G - y_k|_{k \neq G}  = \dfrac{1}{1+\lambda} > 0$, the generated example should always be correctly classified, thereby meeting the first underconfidence criterion. Furthermore, its confidence level should be $y_G$ and negatively correlated with $\lambda$. Since the second condition requires minimizing confidence as much as possible, ideally, $\lambda$ should be maximized. However, considering implementation limitations, we can only approximate the optimization and cannot guarantee reaching it, so a certain margin between $y_G$ and $y_k$ is necessary. 
As $y_G - y_k= \dfrac{1}{1+\lambda}$ is inversely related to $\lambda$, we need to select a smaller $\lambda$ to ensure classification accuracy.
Given the conflicting requirements for $\lambda$, we conduct an experiment, and the results are presented in the next section to identify a trade-off between these two conditions for underconfidence examples.

Finally, by solving~\cref{IAA:eq:advupdate}, we generate the IAEs with the highest potential for revealing the contradiction between classification result and confidence. These examples serves as an augmentation during training to enhance the calibration of the DNN model. 
%
The specified step of generating IAEs is listed in \cref{IAA:alg:IAE}.




\begin{algorithm}
\caption{Inverse Adversarial Attack}\label{IAA:alg:IAE}
\begin{algorithmic}
\State \textbf{Input:}
\State $\X \gets clear~input$
\State $\N_0 \gets random~noise~map$
\State $\G \gets ground~truth~label$
\State $\epsilon \gets maximum~perturbation$
\State $T \gets iteration~number$
\State $t \gets 1$
\State $\alpha \gets \epsilon / T$
\State \textbf{Output}: Inverse adversarial examples

\While{$t \neq T$}
\State $l \gets \mathcal{L}(\theta,\X,\N_{t-1},\G)$
\State $grad \gets \mathrm{sign}(\nabla l_{\N_{t-1}})$
\State $\N_{t} \gets \N_{t-1} ~-~\alpha * grad$
\State $\N_{t} \gets \|\N_{t}\|_{\infty} \leq t*\epsilon / T$
\State $t=t+1$
\EndWhile
\State return $\X+\N_T$
\end{algorithmic}
\end{algorithm}


\begin{figure*}[!th]
\centering
        \centering
        \includegraphics[width=.7\linewidth]{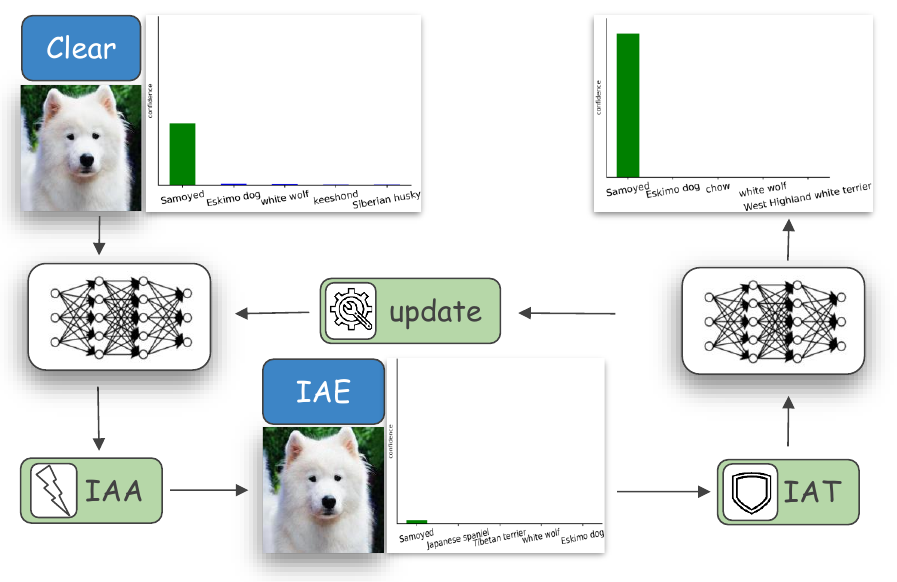}
        \caption{The framework of Inverse Adversarial Training.}    
\label{IAA:fig:framrwork}
\end{figure*}

\subsection{Inverse adversarial Training}
\label{IAA:subsec:IAT}
The existence of IAEs demonstrates that the normally trained model may provide underconfidence prediction results. This may result in an unexpected reject problem (underconfidence) for an over-reliance AI model. For example, a self-driving system may have a delay response in recognizing sign on the expressway.

%
To avoid this phenomenon, we propose an \textit{Inverse Adversarial Training} (IAT), in which we add the IAEs in the training process. They use the capability of these examples to reveal the uncalibrated cases and enhance the model's calibration. The framework of our proposed training is illustrated in \cref{IAA:fig:framrwork}.

Usually, a plain training process is to optimize such a target function:
\begin{align}
\label{IAT:eq:trainplain}
\mathop{\mathrm{min}}\limits_{\theta} \mathbb{E}_{(\X,\G) \sim \mathcal{D}} [\ell(\F(\X), \G)],
\end{align}
where ${\mathcal{D}}$ is the data distribution. The normal training process updates the parameters $\theta$ in the DNN model by minimizing the loss.

The target of IAT is to minimize the loss of uncalibrated examples. Different from Adversarial Training, which focus on increasing the accuracy of overconfident adversarial examples to guarantee the robustness of model, the IAT adjust the outputs of underconfident examples generated by models.
Thus, the objective function~\cref{IAT:eq:trainplain} is then modified with~\cref{IAT:eq:IAE2} for IAT:
\begin{align}
\label{IAT:eq:IAT}
\mathbb{IAT}:~\mathop{\mathrm{min}}\limits_{\theta} \mathbb{E}_{(\X,\G) \sim \mathcal{D}} [\ell(\F(\X+\mathop{\mathrm{argmin}}\limits_{\N \in \Omega}\mathcal{L}), \G)],
\end{align}
where $\Omega$ represents the perturbation space, which is $\|N\| \leq \epsilon$.
During the training phase, IAT incorporates IAEs into the model, addressing underconfidence cases.

In the implementation, IAT updates the model's parameters in $I$ times iteration. We first generate the noisy map $\N_0$ with the initial model $f_{\theta_0}$. Afterward, in each iteration $i$, we update the model's parameters $\theta_i$ according to the gradient descent:
\begin{align}\label{IAT:eq:optim_CAT}
\theta_{i} = \theta_{i-1} ~ - ~ \eta~ \nabla_{\theta} \ell(\theta_{i-1},\X,\N_{i-1},\G),
\end{align}
where $\eta$ refers to the learning rate.

%% file: sec/4_experiments.tex
\section{Experiments}
\label{IAT:sec:exp}

\begin{figure*}[!th]
\centering
        \centering
        \begin{subfigure} {.32\textwidth}
        \includegraphics[width=1\linewidth]{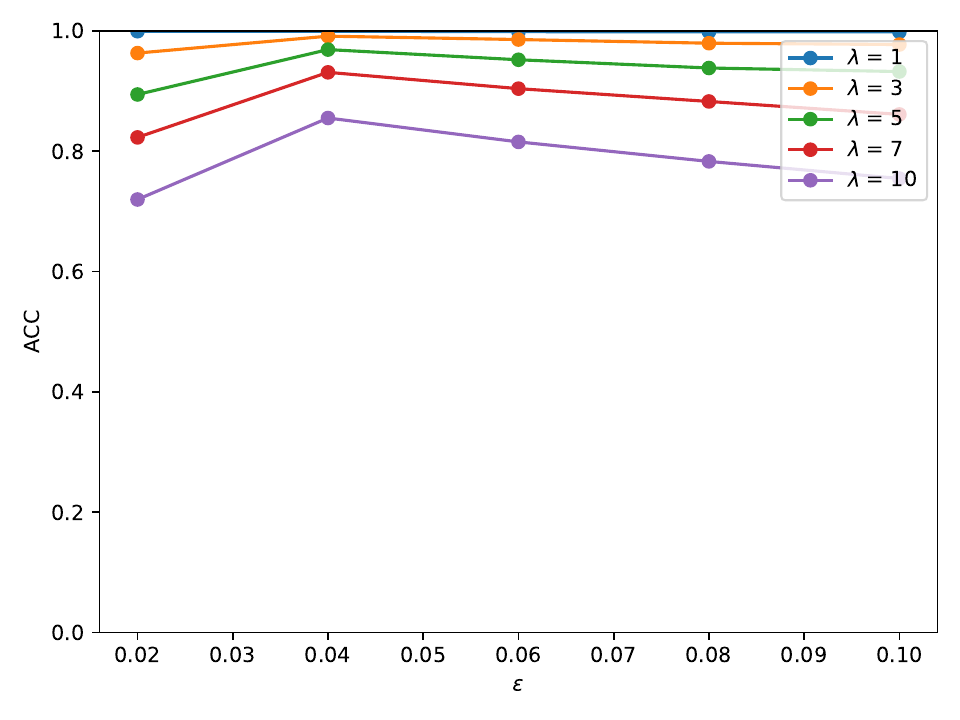}
        \caption{}
        \end{subfigure} 
        \begin{subfigure} {.32\textwidth}
        \includegraphics[width=1\linewidth]{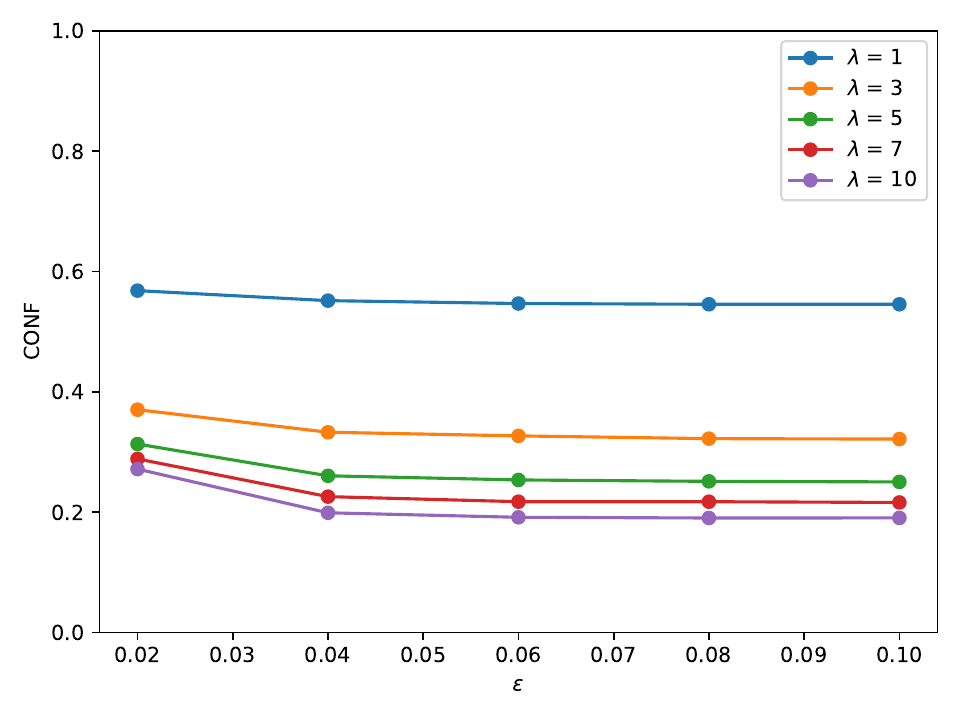}
        \caption{}
        \end{subfigure} 
        \begin{subfigure} {.32\textwidth}
        \includegraphics[width=1\linewidth]{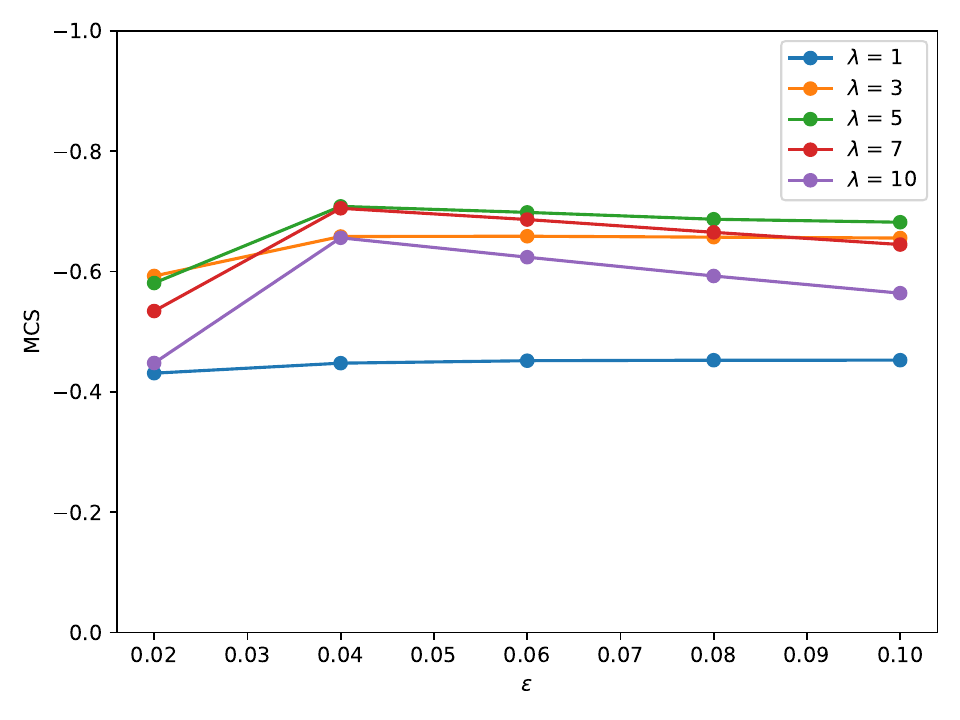}
        \caption{}
        \end{subfigure} 
        \caption{Apply different IAA with different $\lambda$ against to a well-trained model on CIFAR10. ACC, CONF and MCS are shown along with $\epsilon$ ranging from $0.02$ to $0.1$.}    
\label{IAA:fig:lambda_tune}
\end{figure*}



\paragraph{Metric:}
\textbf{Accuracy (ACC)} represents the proportion of correctly predicted cases among all input images. \textbf{Confidence (CONF)} refers to the maximum value of the softmax output in the classifier network's final layer. Note that ``the confidence of the ground truth label'' refers to the probability assigned by a classifier to the ground truth label as its prediction.
As our work focuses more on the underconfidence problem, we choose \textbf{Miscalibration Score (MCS)} \cite{pmlr-v216-ao23a} as our metric to evaluate the calibration of a model during our experiments. Different with the widely-used Expected Calibration Error (ECE) \cite{guo2017calibration} that considers the inconformity of accuracy and confidence equally, MCS expresses overconfidence and underconfidence with the positive or negative value.
MCS is defined by the prediction accuracy and confidence:
\begin{align}
\label{IAT:eq:MCS}
MCS =  \sum_{m=1}^{M} \frac{|B_m|}{N} (acc(B_m)-conf(B_m)),
\end{align}
where $M$ is the number that separates the confidence range into several bins and $N$ indicates the total number of cases. Generally, MCS reflects the inconsistency between accuracy and confidence.
\textbf{Generally, an MCS value approaching $0$ indicates better model calibration. A higher absolute MCS value suggests a stronger ability of the attack to expose a model's miscalibration issues, where positive values indicate overconfidence and negative values indicate underconfidence.}

\paragraph{Datasets and Threaten models:} 


 For most experiments, we use ResNet18 \cite{He2015DeepRL} as the backbone of the classifier. Our proposed IAT and AT+IAT models are evaluated alongside other models using several baseline training strategies, including plain training (PT), PGD adversarial training (PGD-AT) \cite{madry2017towards}, confidence-calibrated adversarial training (CCAT) \cite{stutz2020confidence}, and adversarial calibration entropy-enhanced training (ACE-AT) \cite{chen2024integrating}. It is important to note that the AT strategy applied in AT+IAT follows the PGD-AT method.
Most experiments are conducted on the CIFAR10 dataset \cite{krizhevsky2009learning}, with the exception of a qualitative analysis performed on the NeurIPS'17 adversarial competition dataset \cite{kurakin2018adversarial} using a pretrained VGG19 model \cite{simonyan2014very}.

\subsection{$\lambda$ evaluation}
In this section, we evaluate the ability to reveal underconfidence issues using different values of $\lambda$. As discussed in \cref{IAA:subsec:optim}, larger values of $\lambda$ result in lower confidence, while smaller values help maintain accuracy. Based on this, we apply IAA to a well-trained ResNet18 model on the CIFAR-10 dataset with various $\lambda$ values. The results are shown in \cref{IAA:fig:lambda_tune}. 

We illustrate the ACC, CONF and MCS of IAA with different $\lambda=1, 3, 5, 7, 10$.
MCS is a signed score representing the miscalibration of a model. With a higher absolute value  the model is more miscalibrated.
\cref{IAA:fig:lambda_tune} (a) indicates that the ACC decreases with the increase of $\lambda$, which demonstrates our concern of safe margin for accuracy.
\cref{IAA:fig:lambda_tune} (b) illustrates the negative correlation between $\lambda$ and $y_G$ in minimization point of $\mathcal{L}$, which is consistent with our previous conclusion.
Referring to the MCS values in \cref{IAA:fig:lambda_tune} (c), we choose $\lambda = 5$ for IAA in the following experiments.
As a result, the $y_G=0.25$ with $K=10$ and $y_G=0.1675$ with $K=200$.


\subsection{Underconfidence issues for different backbones}

\begin{figure}[!t]
\centering
        \centering
        \includegraphics[width=.8\linewidth]{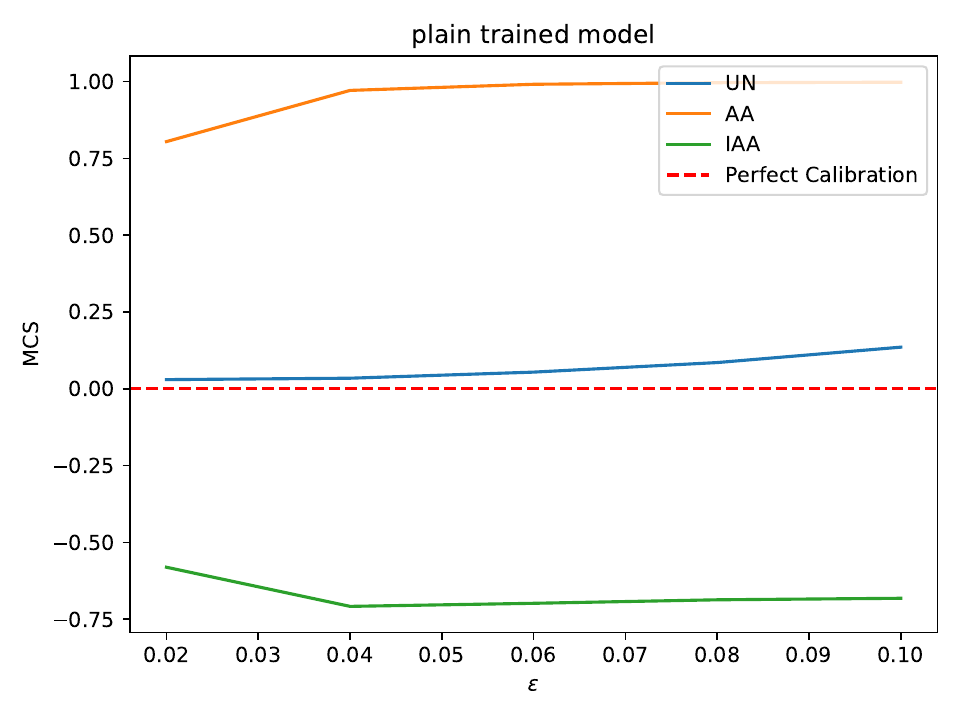}
        \caption{MCS of the examples generated by uniform noise (UN), PGD attack (AA), IAA against to PT model with maximum perturbation $\epsilon$ ranging from $0.02$ to $0.1$. The red dot line indicates the perfect calibration, \ie, MCS=0.}    
\label{IAA:fig:MCS_eps_plain}
\end{figure}

To illustrate the difference in calibration impact between AA and our IAA, we generate uniform noise (UN), PGD adversarial attacks (AA), and IAA with maximum perturbations ranging from $\epsilon = 0.02$ to $0.1$, and plot them along with their MCS. The result is shown in \cref{IAA:fig:MCS_eps_plain}.

It is evident that the MCS of UN increases as $\epsilon$ grows, indicating that the plain trained model exhibits increasing overconfidence with higher levels of perturbation. However, the effectiveness of uniform noise in exposing model miscalibration is limited, as it seldom generates overconfident examples with high MCS.
In contrast, PGD adversarial attack (AA) is capable of generating examples with high MCS even under small perturbations. For instance, AA achieves an MCS of $0.85$ at a perturbation level of $0.02$.

Notably, only IAA is capable of generating examples with negative MCS, thereby revealing the underconfidence issue of the PT model. 
The absolute value of AA exceeds that of IAA, suggesting that achieving overconfidence is easier than inducing underconfidence. 
This can be attributed to the fact that suppressing the softmax confidence for the ground truth label guarantees lower accuracy without conflicting with the increase in maximum softmax confidence. Conversely, reducing the maximum softmax confidence has a pronounced negative impact on accuracy.


\begin{figure}[!th]
\centering
        \centering
        \begin{subfigure} {.4\textwidth}
        \includegraphics[width=1\linewidth]{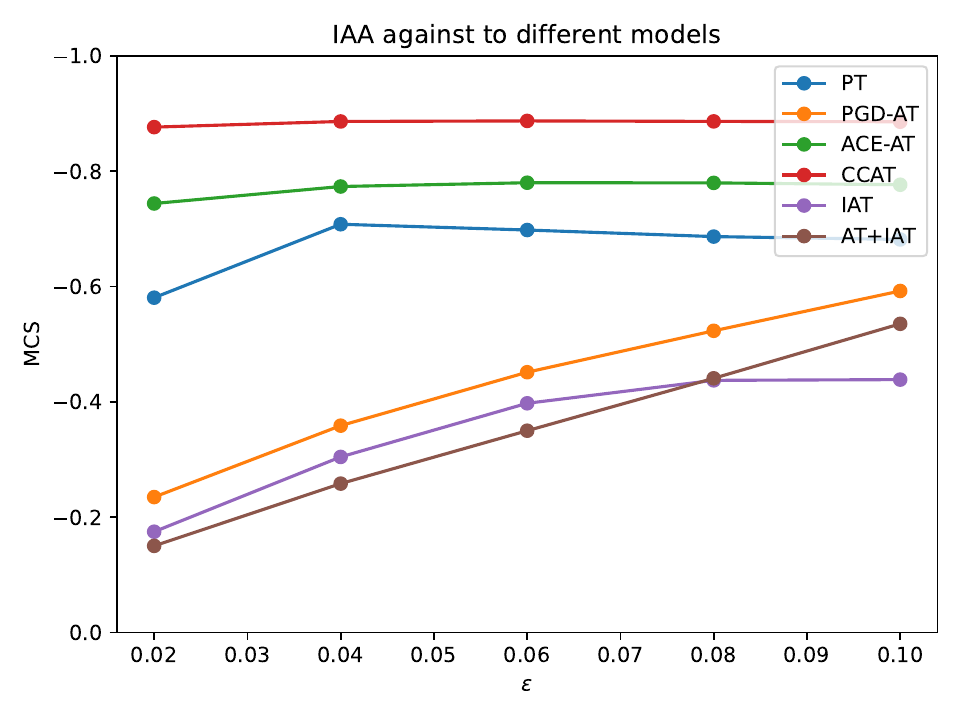}
        \caption{}
        \end{subfigure}
        \begin{subfigure} {.4\textwidth}
        \includegraphics[width=1\linewidth]{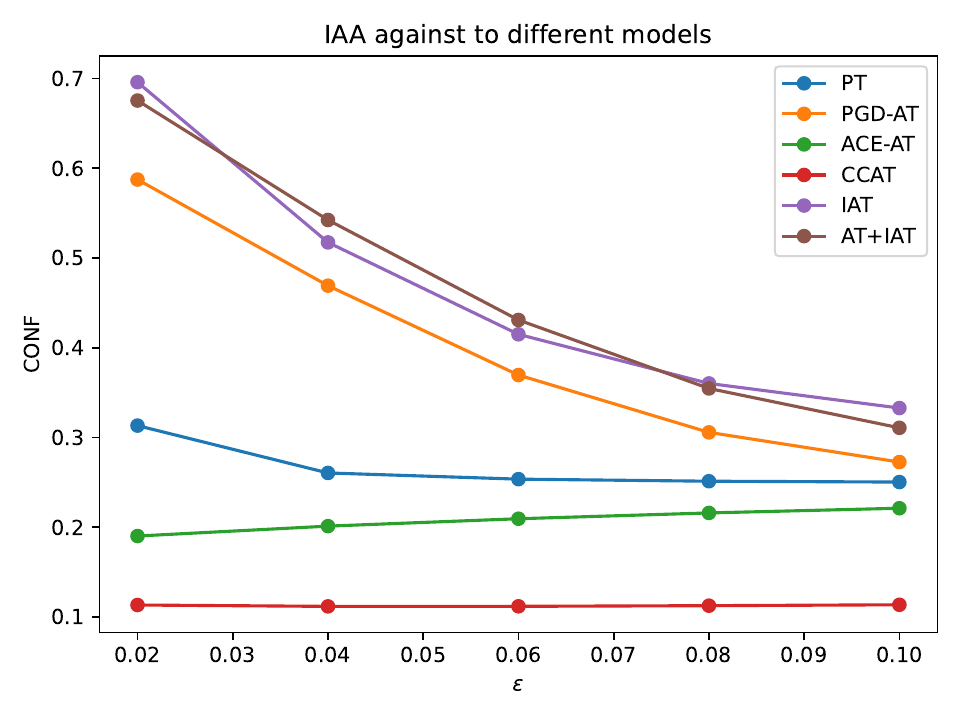}
        \caption{}
        \end{subfigure}
        \caption{MCS and CONF of IAA against to PT, PGD-AT, ACE-AT, CCAT, IAT and AT+IAT models with maximum perturbation $\epsilon$ ranging from $0.02$ to $0.1$.}    
\label{IAA:fig:IAAquanti}
\end{figure}

\begin{table}[!h]
    \centering
    \begin{tabular}{|c|cc|cc|cc|}
        \hline
        \multirow{2}{*}{Models} & \multicolumn{2}{c|}{MCS} & \multicolumn{2}{c|}{ACC} & \multicolumn{2}{c|}{CONF} \\ 
        \cline{2-7}
        & IAA & UCA & IAA & UCA & IAA & UCA \\ 
        \hline
        PT      &  &  &  &  &  &  \\
        PGD-AT  &  &  &  &  &  &  \\
        ACE-AT  &  &  &  &  &  &  \\
        IAT     &  &  &  &  &  &  \\
        AT-IAT  &  &  &  &  &  &  \\
        \hline
    \end{tabular}
    \caption{Performance comparison on CIFAR10 dataset against different models with IAA and UCA of epsilon=0.03.}
    \label{IAA:tab:IAAquanti_cifar10}
\end{table}

\begin{table}[!h]
    \centering
    \begin{tabular}{|c|cc|cc|cc|}
        \hline
        \multirow{2}{*}{Models} & \multicolumn{2}{c|}{MCS} & \multicolumn{2}{c|}{ACC} & \multicolumn{2}{c|}{CONF} \\ 
        \cline{2-7}
        & IAA & UCA & IAA & UCA & IAA & UCA \\ 
        \hline
        PT      &  &  &  &  &  &  \\
        PGD-AT  &  &  &  &  &  &  \\
        ACE-AT  &  &  &  &  &  &  \\
        IAT     &  &  &  &  &  &  \\
        AT-IAT  &  &  &  &  &  &  \\
        \hline
    \end{tabular}
    \caption{Performance comparison on ImageNet Tiny dataset against different models with IAA and UCA of epsilon=0.03.}
    \label{IAA:tab:IAAquanti_imagenettiny}
\end{table}

\begin{figure*}[th]
\centering
        \centering\centering
        \begin{subfigure} {.32\textwidth}
        \includegraphics[width=1\linewidth]{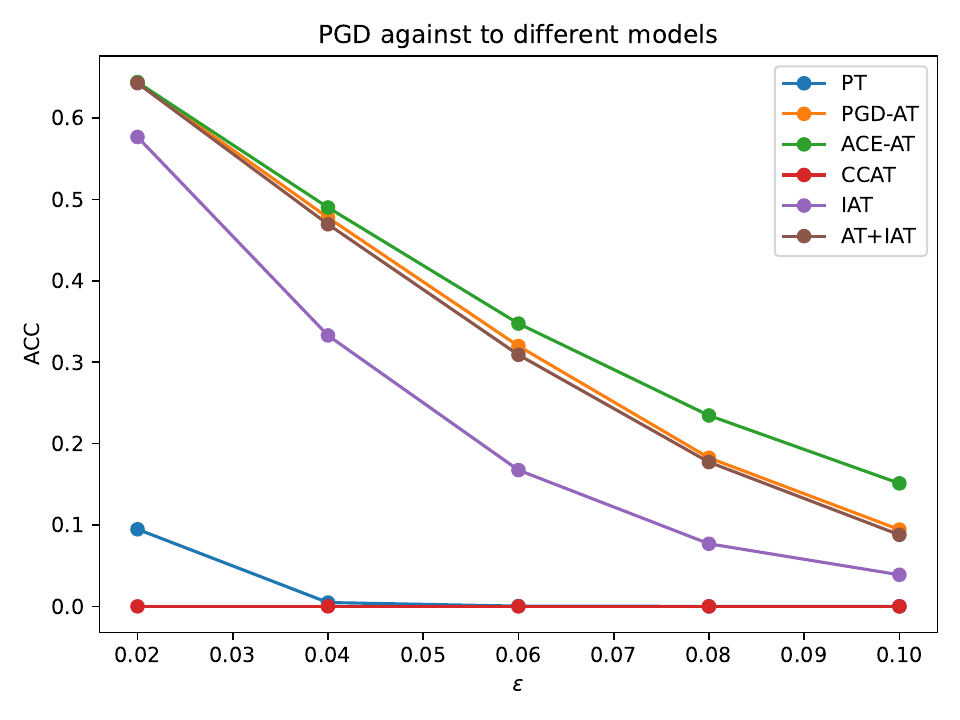}
        \caption{}
        \end{subfigure}
        \begin{subfigure} {.32\textwidth}
        \includegraphics[width=1\linewidth]{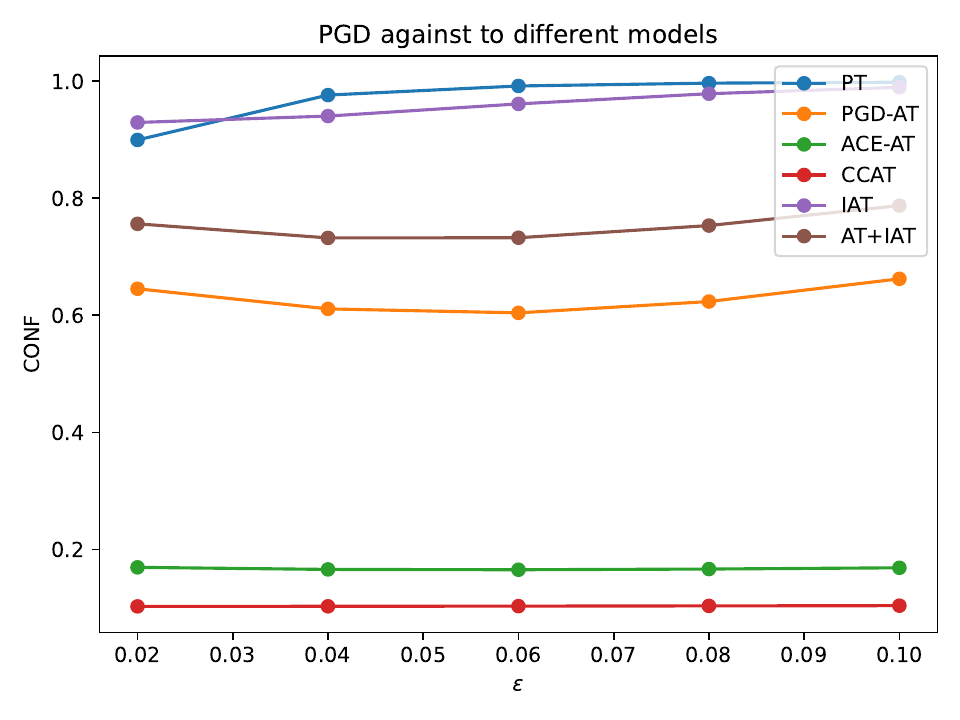}
        \caption{}
        \end{subfigure}
        \begin{subfigure} {.32\textwidth}
        \includegraphics[width=1\linewidth]{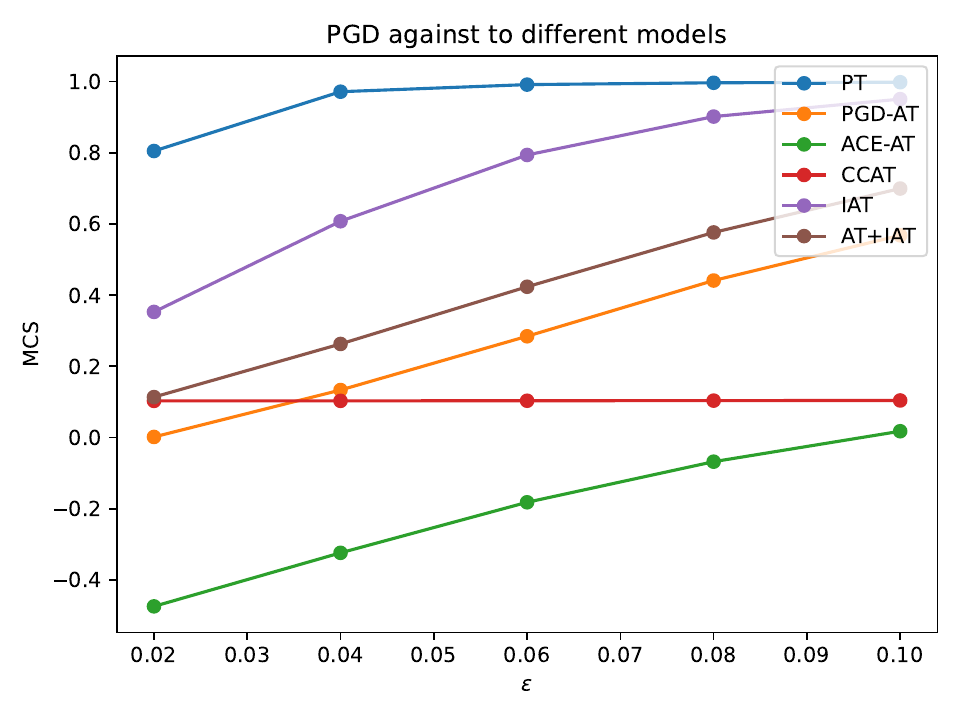}
        \caption{}
        \end{subfigure}
        \caption{ACC, CONF and MCS of PGD attack against to PT, PGD-AT, ACE-AT, CCAT, IAT and AT+IAT models with maximum perturbation $\epsilon$ ranging from $0.02 \sim 0.1$.}    
\label{IAA:fig:PGD}
\end{figure*}

\subsection{Quantitative Analysis of IAA}
\yupeng{add UCA with figure MCS. use the table to illustrate the comparison of IAA and UCA in epsilon=0.03. different models - mcs, acc, conf}
To demonstrate the effectiveness of IAA in exposing underconfidence issues, we evaluate the MCS and CONF of IAEs crafted for models trained with various strategies, including PT, PGD-AT, ACE-AT, CCAT, IAT and AT+IAT.
%
%
Concretely, we vary the $\epsilon$ for different target models to generate multiple MCS-$\epsilon$ curves for clear visual comparison. The results are presented in \cref{IAA:fig:IAAquanti}. 

Generally, the PT model is affected by IAA, while calibrated adversarial training methods (CCAT and ACE-AT) are more sensitive to IAA as they only address the overconfidence issue by minimizing the confidence of any perturbation, as shown in \cref{IAA:fig:IAAquanti} (b).
%
PGD-AT exhibits some resistance to IAA, as it increases the minimum confidence within the $\epsilon$-neighborhood of the input image. Our IAT achieves the best performance in resisting IAA. Furthermore, AT+IAT, which integrates IAEs and AEs into the training process, demonstrates comparable resistance to IAA and even better performance (lower absolute MCS values) at low perturbation levels ($\epsilon < 0.08$).


\subsection{Quantitative Analysis of PGD attack}
In this part, we evaluate the robustness of IAT and AT+IAT in resisting overconfidence attack.
Specifically, we apply the PGD attack against to all baseline models and our proposed IAT and AT+IAT models with a sliding maximum perturbation $\epsilon$ ranging from $0.02$ to $0.1$.
The results are shown in \cref{IAA:fig:PGD}

As shown in \cref{IAA:fig:PGD} (a), although PGD-AT and ACE-AT maintain the highest robustness, IAT can still partially resist PGD (overconfidence) attacks, even as an underconfidence defense method that has not encountered any adversarial examples during training.
Moreover, both PGD-AT and ACE-AT sacrifice confidence to achieve greater robustness, which makes the models more risk-averse.
Besides, with respect to MCS, IAT also shows some resistance compared to PT. PGD-AT demonstrates significant resistance to PGD attacks since it is exposed to PGD examples during training. Note that ACE-AT exhibits an underconfidence phenomenon, as it minimizes confidence to an extremely low level. CCAT achieves the best MCS values (approaching $0$) by minimizing confidence under varying levels of noise, regardless of their accuracy.
Finally, AT+IAT achieves the same level of robustness as PGD-AT while also demonstrating good calibration under IAA. Its strong performance in resisting IAA suggests that combining AT and IAT is an effective approach for producing a well-calibrated model capable of withstanding both overconfidence attacks (\eg, AA) and underconfidence attacks (\eg, IAA).

\begin{figure*}[!th]
\centering
        \centering
        \includegraphics[width=.9\linewidth]{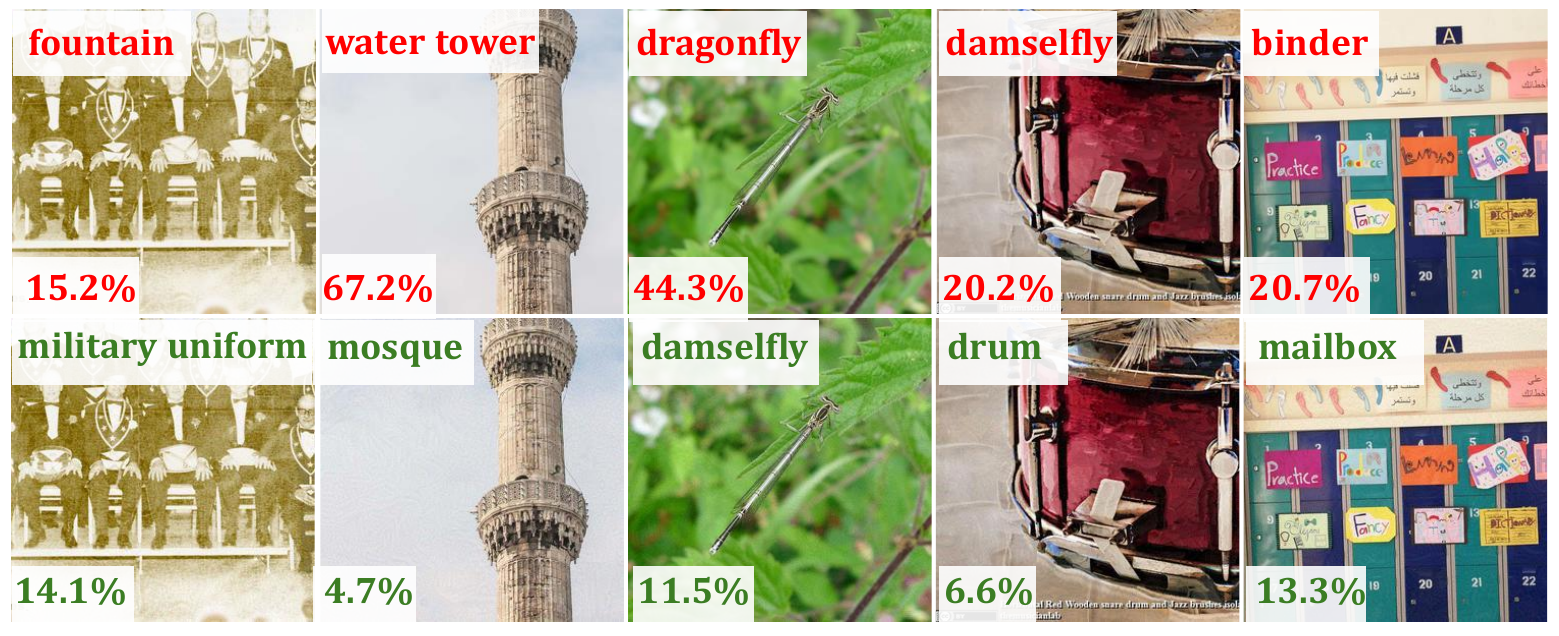}
        \caption{Visualization of inverse adversarial examples generated for the VGG19 network pretrained on ImageNet. Clear input images (wrongly predicted by the pretrained model) are shown in the first row, with IAEs displayed in the second row. Each image displays its classification result from VGG19 in the top-left corner. The predicted confidence levels are shown at the bottom. Incorrect predictions are highlighted in red and correct predictions are highlighted in green.}    
\label{IAA:fig:visual}
\end{figure*}


\subsection{Qualitative Analysis of IAA}
To visually demonstrate the effectiveness of IAA, we conduct an experiment on a high-resolution image dataset from the NeurIPS'17 adversarial competition, specifically the DEV dataset \cite{kurakin2018adversarial}, which is compatible with ImageNet \cite{deng2009imagenet}. We use a VGG19 network pretrained on ImageNet as the threaten model. In this experiment, we set the maximum IAA perturbation $\epsilon$ to $0.03$. The clear input images and corresponding IAEs are shown in \cref{IAA:fig:visual}.

Given that the 1000-class classification task is highly challenging, the top-1 accuracy of this pretrained VGG19 on the NeurIPS2017 dataset only reaches $87.6\%$, resulting in some misclassified cases. For example, the image in the second column is incorrectly predicted as 'water tower' with a high confidence of $67.2\%$, indicating overconfidence in this instance
The reason could be that, from this angle, the mosque indeed bears some resemblance to a water tower. Similar situations also occur in the case shown in the third column.
However, IAA successfully generates examples with correct predictions but lower confidence (e.g., $4.7\%$ for the ambiguous second case) using imperceptible perturbations. This demonstrates that our proposed IAA effectively reveals potential underconfidence issues even in cases prone to confusion.


\subsection{Discussion on Transferability of IAA}

\begin{table}[h!]
\centering

\resizebox{0.45\textwidth}{!}{
\begin{tabular}{|c|c|c|c|}
    \hline
    \diagbox{Target}{Threaten} & ResNet18 & MobileNet & VGG19 \\ \hline
    ResNet18 & -0.6887 & 0.0234 & 0.0313 \\ \hline
    MobileNet & 0.0703 & -0.7192 & 0.0415 \\ \hline
    VGG19 & 0.0664 & 0.0153 & -0.2395 \\ \hline
\end{tabular}
}
\caption{We present the MCS results of transfer and white-box inverse adversarial attacks across three fine-tuned models: ResNet18, MobileNet, and VGG19. The whitebox attack results are shown along the diagonal, while the remaining results correspond to transfer attacks.}
\label{IAA:tab:transferability}
\end{table}

Transferability refers to the ability of an attack to successfully affect a target model using examples generated from a different threaten model. Evaluating the transferability of IAA is important, as it aligns more closely with real-world attack scenarios.
To investigate the transferability of IAA, we apply IAA on three well-trained threaten models, including ResNet18, MobileNet and VGG19, to generate corresponding IAEs with a maximum perturbation of $\epsilon=0.03$. These IAEs are then fed back to the models as target models, yielding nine MCS values that reflect the impact of IAA on each.
Following terminology from the adversarial attack field, when the threaten model and target model are identical, we refer to this as a ``whitebox'' IAA, indicating full access to the target model. Conversely, when the threaten model and target model differ, we refer to it as ``transfer'' IAA.

\cref{IAA:tab:transferability} presents the MCS results for transfer and whitebox IAAs across three fine-tuned models: ResNet18, MobileNet, and VGG19. In the table we have the following observation:

Diagonal entries (\eg, ResNet18-ResNet18) represent the whitebox IAA results, indicating the attack performance on each model’s own architecture is significant.
Specifically, the whitebox IAA against to ResNet18 has a MCS value of $-0.6887$, showing a strong reduction in model calibration under attack.
Same phenonmenon exists to the MobileNet.
The MCS value of whitebox IAA against to VGG19 is $-0.2395$, which, though higher (lower in absolute value) than that of ResNet18 and MobileNet, still shows a substantial impact on calibration by whitebox IAA.

Off-diagonal entries represent transfer attack results. The findings indicate that changing the model architecture significantly affects the effectiveness of IAA. For instance, IAEs crafted from VGG19 yield MCS values of $0.0664$ and $0.0153$ when attacking ResNet18 and MobileNet, respectively.

In summary, whitebox IAA typically produces large negative MCS values, highlighting its strong ability to expose underconfidence issues. In contrast, transfer IAA generally yields MCS values closer to zero or even positive, indicating lower effectiveness. Enhancing the transferability of IAA presents a promising direction for future research.

%% file: sec/5_conclusion.tex
\section{Conclusion}
\label{IAT:sec:conclusion}
Calibration is a crucial feature for reliable AI systems. Recent work has primarily focused on addressing the issue of overconfidence, such as through adversarial attacks. In this paper, we explore the opposite side of miscalibration by proposing a novel task that generates correctly predicted examples with minimal confidence, termed the Inverse Adversarial Attack (IAA). To satisfy the two conditions of underconfidence—correct prediction and low confidence—we incorporate two cross-entropy loss terms to balance the trade-off in generating inverse adversarial examples. We then provide a mathematical analysis of the proposed loss function, demonstrating that its minimization meets the two conditions of underconfidence.
Extensive experimental results show that our IAA effectively reveals underconfidence issues and can be embedded in the training process, forming what we term Inverse Adversarial Training (IAT) to counter such attacks. Combining adversarial training (AT) with IAT provides a practical approach to achieve models that are both robust and well-calibrated.
Additionally, the discussion on the transferability of IAA provides a potential direction for future work.